\pdfoutput=1

\documentclass[11pt]{article}

\usepackage[]{acl}

\usepackage{times}
\usepackage{latexsym}
\usepackage{graphicx}
\usepackage{algpseudocode,algorithm,algorithmicx}
\newcommand*\Let[2]{\State #1 $\gets$ #2}
\usepackage{caption}
\usepackage{subcaption}
\usepackage{booktabs}
\usepackage{sidecap}
\usepackage{hyperref}
\usepackage{multirow}
\usepackage{amsmath,microtype}
\usepackage{tikz}
\usepackage{enumitem}

\usepackage[T1]{fontenc}

\usepackage[utf8]{inputenc}

\usepackage{microtype}

\newcommand{\cgrad}[5]{%
	\ifdim #1 pt > #3 pt
		\pgfmathsetmacro{\PercentColor}{max(min((100.0*(#1-#3)/(#4-#3)),100.0),0.00)} %
		\hspace{-4.3pt}\colorbox{green!\PercentColor!yellow!40!white}{\makebox[#5]{\strut\textcolor{black}{#1}}}\hspace{-4.3pt}
	\else
		\pgfmathsetmacro{\PercentColor}{max(min((100.0*(#3-#1)/(#3-#2)),100.0),0.00)} %
		\hspace{-4.3pt}\colorbox{red!\PercentColor!yellow!40!white}{\makebox[#5]{\strut\textcolor{black}{#1}}}\hspace{-4.3pt}
	\fi
}

\newcommand{\cimtop}[1]{\cgrad{#1}{95.1}{95.3}{95.5}{18pt}}
\newcommand{\csmtop}[1]{\cgrad{#1}{79.1}{80.7}{82.3}{18pt}}
\newcommand{\cimtod}[1]{\cgrad{#1}{97.1}{97.95}{98.8}{18pt}}
\newcommand{\csmtod}[1]{\cgrad{#1}{73.5}{76.1}{79.7}{18pt}}
\newcommand{\cimatis}[1]{\cgrad{#1}{96.0}{96.3}{96.6}{18pt}}
\newcommand{\csmatis}[1]{\cgrad{#1}{75.7}{79.4}{83.1}{18pt}}
\newcommand{\cimean}[1]{\cgrad{#1}{95.9}{96.2}{96.5}{18pt}}
\newcommand{\csmean}[1]{\cgrad{#1}{78.2}{80.175}{82.1}{18pt}}
\newcommand{\covral}[1]{\cgrad{#1}{87.2}{88.25}{89.3}{30pt}}

\newcommand{\scimean}[1]{\cgrad{#1}{95.9}{96.2}{96.5}{36pt}}
\newcommand{\scsmean}[1]{\cgrad{#1}{78.2}{80.175}{82.1}{36pt}}
\newcommand{\scovral}[1]{\cgrad{#1}{87.2}{88.25}{89.3}{45pt}}

\title{CrossAligner \& Co: Zero-Shot Transfer Methods for Task-Oriented Cross-lingual Natural Language Understanding}

\author{Milan Gritta\textsuperscript{1,$\dagger$}, Ruoyu Hu\textsuperscript{2,$\dagger$,}\thanks{\ \ Work conducted as Research Intern at Huawei's Noah's Ark Lab, London. $\dagger$ - Equal contribution.} {\normalfont\ and} Ignacio Iacobacci\textsuperscript{1}\\
	\textsuperscript{1}Huawei Noah’s Ark Lab, London, UK\\
	\textsuperscript{2}Imperial College London, UK \\
	\texttt{\{milan.gritta,ignacio.iacobacci\}@huawei.com} \\
	\texttt{ruoyu.hu18@imperial.ac.uk}}

\begin{document}
\maketitle
\begin{abstract}

Task-oriented personal assistants enable people to interact with a host of devices and services using natural language. One of the challenges of making neural dialogue systems available to more users is the lack of training data for all but a few languages. Zero-shot methods try to solve this issue by acquiring task knowledge in a high-resource language such as English with the aim of transferring it to the low-resource language(s). To this end, we introduce \textbf{CrossAligner}, the principal method of a variety of effective approaches for zero-shot cross-lingual transfer based on learning alignment from unlabelled parallel data. We present a quantitative analysis of individual methods as well as their weighted combinations, several of which exceed state-of-the-art (SOTA) scores as evaluated across nine languages, fifteen test sets and three benchmark multilingual datasets. A detailed qualitative error analysis of the best methods shows that our fine-tuned language models can zero-shot transfer the task knowledge better than anticipated.

\end{abstract}

\section{Introduction}

Natural language understanding (NLU) refers to the ability of a system to `comprehend' the meaning (semantics) and the structure (syntax) of human language \cite{wang2019superglue} to enable the interaction with a system or device. Cross-lingual natural language understanding (XNLU) alludes to a system that is able to handle multiple languages simultaneously \cite{artetxe2019massively,hu2020xtreme}. We focus on task-oriented XNLU that comprises two correlated objectives: i) Intent Classification, which identifies the type of user command, e.g. `edit\_reminder', `send\_message' or `play\_music' and ii) Entity/Slot Recognition, which identifies relevant entities in the utterance including their types such as dates, messages, music tracks, locations, etc. In a modular dialogue system, this information is used by the dialogue manager to decide how to respond to the user \cite{casanueva2017benchmarking,gritta2021conversation}. For neural XNLU systems, the limited availability of annotated data is a significant barrier to scaling dialogue systems to more users \cite{razumovskaia2021crossing}. Therefore, we can use cross-lingual methods to zero-shot transfer the knowledge learnt in a high-resource language such as English to the target language of choice \cite{artetxe2019cross,siddhant2020evaluating}. To this end, we introduce a variety of alignment methods for zero-shot cross-lingual transfer, most notably \textbf{CrossAligner}. Our methods leverage unlabelled parallel data and can be easily integrated on top of a pretrained language model, referred to as \textbf{XLM}\footnote{Not to be confused with \newcite{lample2019cross}.}, such as XLM-RoBERTa \cite{conneau2019unsupervised}. Our methods help the XLM align its cross-lingual representations while optimising the primary XNLU tasks, which are learned only in the source language and transferred zero-shot to the target language. Finally, we also investigate the effectiveness of simple and weighted combinations of multiple alignment losses, which leads to further model improvements and insights. Our contributions are summarised as follows:

\vspace{-0.1cm}
\begin{itemize}[itemsep=1mm, parsep=0pt]
	\setlength\itemsep{0.25em}
	\item We introduce CrossAligner, a cross-lingual transfer method that achieves SOTA performance on three benchmark XNLU datasets.
	\item We introduce Translate-Intent, a simple and effective baseline, which outperforms its commonly used counterpart `Translate-Train'.
	\item We introduce Contrastive Alignment, an auxiliary loss that leverages contrastive learning at a much smaller scale than past work.
	\item We introduce weighted combinations of the above losses to further improve SOTA scores.
	\item Qualitative analysis aims to guide future research by examining the remaining errors.
\end{itemize}

\section{Related Work}
\label{related}

Several approaches to zero-shot cross-lingual transfer exist and can broadly be divided into: a) \textit{Data-based Transfer}, which focuses on training data transformation and b) \textit{Model-based Transfer} that centres around modifying models' training routine. 

\paragraph{Data-based Transfer} Translating utterances for the intent classification task is relatively straightforward so previous works focused on projecting and/or aligning the entity labels between translated utterances. This is followed by standard supervised training with those pseudo-labels and is commonly known as the \textit{translate-train} method. One of the earliest works still being used for this purpose is \textit{fastalign} \cite{dyer2013simple}. It's an unsupervised word aligner trained on a parallel corpus to map each word (thus its entity label) in the source utterance to the word(s) in the target user utterance. Projecting the entity labels can also be done with word-by-word translation and source label copying \cite{yi2021zero}. A teacher model then weakly labels the target data, which is used to train the final student model. Sometimes, this type of label projection is complemented with an additional entity alignment step \cite{li2021cross}. Better performance can be achieved by using machine translation with entity matching and distributional statistics \cite{jain2019entity} though this can be a costly process for each language. A category of `word substitution' methods such as code-switching \cite{qin2020cosda,kuwanto2021low} or dictionary-enhanced pretraining \cite{chaudhary2020dict} have also been shown to improve cross-lingual transfer. 

\paragraph{Model-based Transfer} Prior to the adoption of multilingual transformers \cite{lample2019cross}, task-oriented XNLU methods employed a BiLSTM encoder combined with different multilingual embeddings \cite{schuster2018cross}. Newer approaches usually involve a pretrained XLM and the addition of some new training component(s) with the inference routine remaining mostly unchanged. \citet{xu2020end} learn to jointly align and predict entity labels by fusing the source and target language embeddings with attention and using the resulting cross-lingual representation for entity prediction. \citet{qi2020translation} include an adversarial language detector in training whose loss encourages the model to generate language-agnostic sentence representations for improved zero-shot transfer. \citet{pan2020multilingual} and \citet{chi2020infoxlm} added a contrastive loss to pretraining that treats translated sentences as positive examples and unrelated sentences as negative samples. This training step helps the XLM produce similar embeddings in different languages. However, these methods require large annotated datasets and expensive model pretraining \cite{chi2020infoxlm}. Our proposed methods only use the English task data (which is relatively limited) and its translations for each language.
\vspace{0.2cm}

The most related prior works are \citet{arivazhagan2019missing} for machine translation and \citet{gritta2021xeroalign} for task-oriented XNLU. Both of these are cross-lingual alignment methods that use translated training data to zero-shot transfer the source language model to the target language. We focus on the latter work, called XeroAlign, which reported the most recent SOTA scores on our evaluation datasets. XeroAlign works by generating a sentence embedding of the user utterance for each language, e.g. English (source) and Thai (target) using the $\mathsf{CLS}$ token of the XLM. A Mean Squared Error loss function minimises the difference between the multilingual sentence embeddings and is backpropagated along with the main task loss. XeroAlign aims to bring sentence embeddings in different languages closer together with a bias towards intent classification due to the $\mathsf{CLS}$ embedding, which is the standard input to the intent classifier. We reproduce this method for analysis and comparisons but add a small post-processing step that distinctly improves the reported scores. 

\section{Methodology}
\label{method}

\subsection{CrossAligner}
\label{crossaligner}

\paragraph{Intuition}

We introduce CrossAligner, the most notable of our proposed cross-lingual alignment methods, outlined in Algorithm \ref{alg:crossaligner}. CrossAligner enables effective zero-shot transfer by leveraging unlabelled parallel data for our new language-agnostic objective created through a transformation of the English entity labels. CrossAligner was borne out of early error analysis where we observed that the model incorrectly predicted entities that didn't occur in the input and failed to predict entities that did occur in the input. Using this insight as our main motivation, the essence of CrossAligner is being able to exploit information about the \textit{presence of entities/slots in the user utterance}.

\paragraph{Algorithm} We have used a proprietary service (Huawei Translate) to translate the English user utterances $\mathsf{X_{Eng}}$ into each target language $\mathsf{X_{Tar}}$, however, a publicly available translator can also be used. Note that we use the same translations for each of our alignment methods to compare them fairly.
Our language-agnostic objective is created by transforming the English slot labels $\mathsf{y_{ec}}$ into a fixed binary vector $\mathsf{y_{ca}}$ indicating which entities are present in the input (lines 1-7 in Algorithm \ref{alg:crossaligner}), irrespective of the frequency of their occurrence. 
\vspace{0.2cm}

The standard XNLU training (lines 15-20) features an Intent Classifier ($\mathsf{IC}$) and an Entity Classifier $(\mathsf{EC})$. Each computes a cross-entropy loss ($\mathsf{ce\_loss}$) with a softmax activation using English labelled data (multi-class classification). This yields the standard losses $\mathsf{\mathcal{L}_{ic}}$ and $\mathsf{\mathcal{L}_{ec}}$. The CrossAligner ($\mathsf{CA}$) classifier then pools the $\mathsf{EC}$ logits matrix by reshaping it into a long vector (lines 24 and 29) and predicts which entities are present in the user utterance (multi-label classification). We compute a Binary Cross-Entropy loss ($\mathsf{bce\_loss}$) with a sigmoid activation between the predicted labels $\mathsf{pred_{eng}}$ and $\mathsf{pred_{tar}}$ (for English and Target languages respectively) and our language-agnostic labels $\mathsf{y_{ca}}$ (lines 26 and 31). This yields the CrossAligner losses $\mathsf{\mathcal{L}_{eng}}$ and $\mathsf{\mathcal{L}_{tar}}$. The fact that these gradients are propagated through the $\mathsf{EC}$ to the $\mathsf{XLM}$ token embeddings ensures a good alignment for entity/slot recognition, as shown in the results section. Note that $\mathsf{EC}$, $\mathsf{IC}$ and $\mathsf{CA}$ are shared between languages to aid zero-shot cross-lingual transfer. 

\paragraph{BIO versus IO}
\label{bio}
Using the BIO sequence tagging format \cite{sang2003introduction} can introduce easily avoidable model errors, e.g. predicting a B-tag after an I-tag, two B-tags in succession or skipping the B-tag altogether. We have therefore simplified the training process by making it agnostic w.r.t. the entity's BI order. The B-tags were removed in preprocessing, meaning the entity classifier predicts only IO-tags. At inference, the B-tags get restored with a simple post-processing rule. Note that all our models use this IO-only training.

\begin{figure}[h!]
	\centering
	\includegraphics[width=\linewidth]{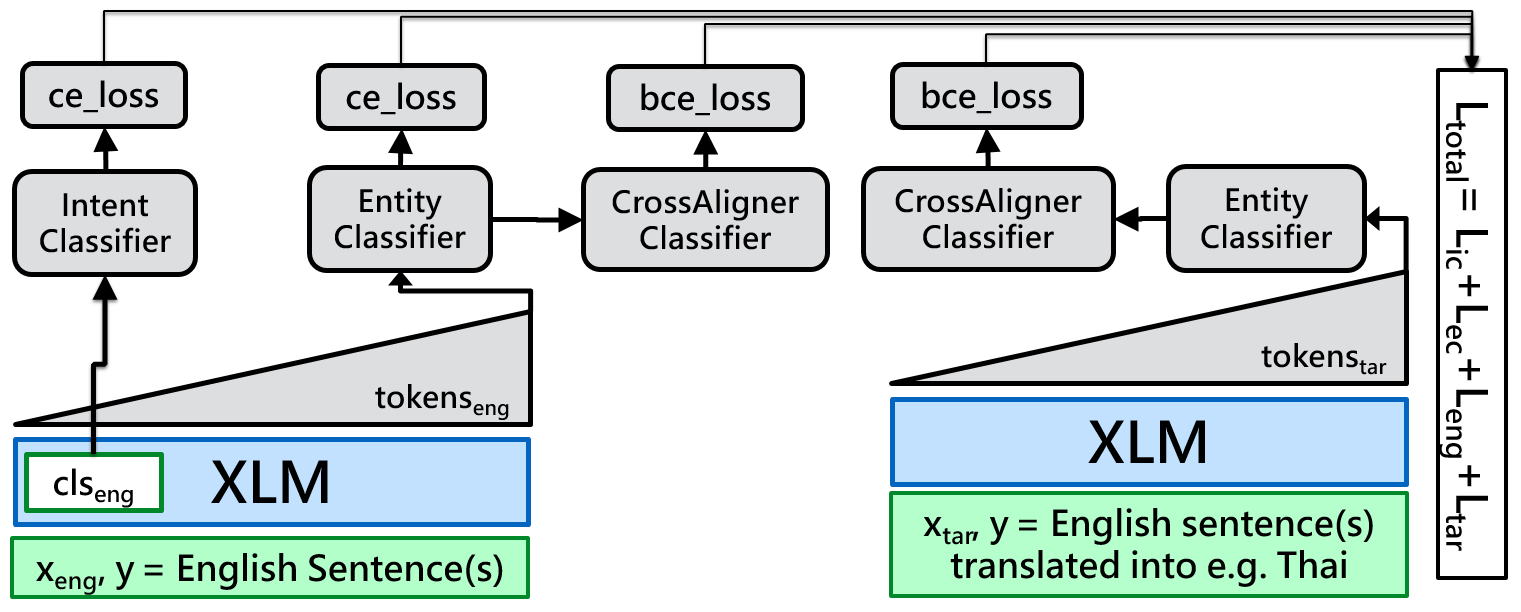}
	\caption{The architecture of CrossAligner. The parameters of the XLM model and all classifiers are shared between languages to enable cross-lingual transfer.}
	\label{fig:xeroaligner}
\end{figure}

\paragraph{Architecture} We use a common task-oriented XNLU model that employs a pretrained $\mathsf{XLM}$, e.g. JointBERT \cite{chen2019bert}. The $\mathsf{IC}$, $\mathsf{EC}$ and $\mathsf{CA}$ each feature a single multi-layer perceptron of sizes: $\mathsf{[hidden\_size, len(intent\_classes)]}$, $\mathsf{[hidden\_size, len(entity\_classes)]}$ and $\mathsf{[seq\_len \times}$ $\mathsf{len(entity\_classes), len(entity\_classes)]}$. Depending on the dataset, $\mathsf{seq\_len}$ varies between 50-100 tokens. The model architecture is shown in Fig \ref{fig:xeroaligner}. 

\setlength{\textfloatsep}{18pt}
\begin{algorithm}[t!]
	\caption{The CrossAligner alignment/loss.}
	\begin{algorithmic}[1]
		\Function{TransformLabels}{$\mathsf{y_{ec}}$}
		\Let{$\mathsf{y_{ca}}$}{$\mathsf{zeros(len(entity\_classes))}$}
		\For{$\mathsf{entity \in y_{ec}}$}
		\Let{$\mathsf{y_{ca}[index\_of(entity)]}$}{$\mathsf{1}$}
		\EndFor
		\State \Return{$\mathsf{y_{ca}}$}
		\EndFunction
		\Statex
		\Let{$\mathsf{XLM}$}{Cross-lingual language model}
		\Let{$\mathsf{IC}$}{Intent Classifier}
		\Let{$\mathsf{EC}$}{Entity Classifier}
		\Let{$\mathsf{CA}$}{CrossAligner Classifier}
		\Let{$\mathsf{X_{Eng}}$}{Standard training data in English}
		\Let{$\mathsf{X_{Tar}}$}{$\mathsf{X_{Eng}}$ translated into Target language}
		\Statex
		\For{$(\mathsf{x_{eng}}, \mathsf{y}), (\mathsf{x_{tar}}, \mathsf{y}) \in \mathsf{X_{Eng}, X_{Tar}}$}
		\Statex{\ \ \ \ \ \ \ \ \small{---Standard XNLU Training---}}
		\Let{$\mathsf{y_{ic}, y_{ec}}$}{$\mathsf{y}$}
		\Let{$\mathsf{cls_{eng}}, \mathsf{tokens_{eng}}$}{$\mathsf{XLM}(\mathsf{x_{eng}})$}
		\Let{$\mathsf{pred_{ic}}$}{$\mathsf{IC}(\mathsf{cls_{eng}})$}
		\Let{$\mathsf{\mathcal{L}_{ic}}$}{$\mathsf{ce\_loss(pred_{ic}, y_{ic})}$}
		\Let{$\mathsf{pred_{ec}}$}{$\mathsf{EC(tokens_{eng})}$}
		\Let{$\mathsf{\mathcal{L}_{ec}}$}{$\mathsf{ce\_loss(pred_{ec}, y_{ec})}$}
		\Statex
		\Statex{\ \ \ \ \ \ \ \ \small{---CrossAligner Training---}}
		\Let{$\mathsf{y_{ca}}$}{$\Call{TransformLabels}{\mathsf{y_{ec}}}$}
		\Let{$\mathsf{shape}$}{$\mathsf{(seq\_len \times len(entity\_classes))}$}
		\Let{$\mathsf{logits_{eng}}$}{$\mathsf{EC(tokens_{eng})}$}
		\State $\mathsf{logits_{eng}.reshape\_matrix\_into(shape}$)
		\Let{$\mathsf{pred_{eng}}$}{$\mathsf{CA(logits_{eng})}$}
		\Let{$\mathsf{\mathcal{L}_{eng}}$}{$\mathsf{bce\_loss(pred_{eng}, y_{ca})}$}
		\Let{$\mathsf{cls_{tar}, tokens_{tar}}$}{$\mathsf{XLM(x_{tar})}$}
		\Let{$\mathsf{logits_{tar}}$}{$\mathsf{EC(tokens_{tar})}$}
		\State $\mathsf{logits_{tar}.reshape\_matrix\_into(shape)}$)
		\Let{$\mathsf{pred_{tar}}$}{$\mathsf{CA(logits_{tar})}$}
		\Let{$\mathsf{\mathcal{L}_{tar}}$}{$\mathsf{bce\_loss(pred_{tar}, y_{ca})}$}
		\Let{$\mathsf{\mathcal{L}_{total}}$}{$\mathsf{\mathcal{L}_{ic} + \mathcal{L}_{ec} + \mathcal{L}_{eng} + \mathcal{L}_{tar}}$}
		\EndFor
	\end{algorithmic}
	\label{alg:crossaligner}
	\vspace{0.15cm}
\end{algorithm}

\subsection{Contrastive Alignment for XNLU}

Our contrastive alignment is based on InfoNCE \cite{vandenoord2018rep}. Previous work has employed a contrastive loss for cross-lingual alignment \cite{pan2020multilingual}, however, the datasets were out-of-domain and orders of magnitude larger. We show that strong results can be obtained using only in-domain (fine-tuning) data. Similar to \cite{wu2021rethink}, if given a randomly sampled batch of $\mathsf{N}$ English sentences $\mathsf{X_{Eng}}$ and its parallel sentences $\mathsf{X_{Tar}}$ in the target language, then the loss on the $\mathsf{i^{th}}$ sentence pair $\mathsf{x_{eng_i} \in X_{Eng}}$ and $\mathsf{x_{tar_i} \in X_{Tar}}$ equals:

\vspace{-0.1cm}
\begin{align}
 \mathsf{\ell (x_{eng_i}, x_{tar_i}) = - \log\frac{e^{sim(x_{eng_i}, x_{tar_i})}}{\sum_{k=1}^{N}e^{sim(x_{eng_i}, x_{tar_k})}}}
 \raisetag{45pt}
\end{align}

\noindent where $\mathsf{sim(u, v) = u \cdot v\ /\ ||u||_2 \cdot ||v||_2}$ is the cosine similarity between two sentence embeddings. A sentence $\mathsf{x_{eng_i} \in X_{Eng}}$ symmetrically forms a positive pair with its translation $\mathsf{x_{tar_i} \in X_{Tar}}$ while the other $\mathsf{N - 1}$ sentence embeddings are treated as negative samples. The batch loss is calculated as the average of all positive pair losses. Algorithm \ref{alg:contrastive} below shows the steps that replace/complement the CrossAligner block (lines 21-32 in Algorithm \ref{alg:crossaligner}).

\begin{algorithm}[h]
	\caption{The Contrastive Alignment loss.}
	\begin{algorithmic}[1]
	\Let{$\mathsf{cls_{eng}}, \mathsf{tokens_{eng}}$}{$\mathsf{XLM}(\mathsf{x_{eng}})$}
	\Let{$\mathsf{cls_{tar}}, \mathsf{tokens_{tar}}$}{$\mathsf{XLM}(\mathsf{x_{tar}})$}
	\Let{$\mathsf{sim}$}{$\mathsf{batch\_cosine\_sim(cls_{eng}, cls_{tar})}$}
	\Let{$\mathsf{labels}$}{$\mathsf{arange(batch\_size)}$}
	\Let{$\mathsf{\mathcal{L}_{cl}}$}{$\mathsf{ce\_loss(sim, labels)}$}
	\Let{$\mathsf{\mathcal{L}_{total}}$}{$\mathsf{\mathcal{L}_{ic} + \mathcal{L}_{ec} + \mathcal{L}_{cl}}$}
    \end{algorithmic}
    \label{alg:contrastive}
\end{algorithm}

\subsection{Translate-Intent}

The translate-train method is used in multilingual NLP as a competitive baseline \cite{liang2020xglue,hu2020xtreme}. After machine translation, the sequence tagging tasks require an additional transformation, i.e. entity label projection and/or word alignment \cite{schuster2018cross,li2020mtop,xu2020end}. This is followed by supervised fine-tuning with the new pseudo-labels. However, both label projection and word alignment are \textit{sources of common errors}. We therefore introduce a simpler baseline called Translate-Intent, which to the best of our knowledge, has not been featured in task-oriented XNLU. We omit the entity/slot recognition for the target language (given the unreliable pseudo-labels) and only use the $\mathsf{IC}$, which is trained with the parallel data $\mathsf{X_{Tar}}$ (labels copied from English). Algorithm \ref{alg:translate} above shows the steps that either replace or complement (in case of a combination of multiple losses) the CrossAligner steps, shown in lines 21-32 in Algorithm \ref{alg:crossaligner}.

\begin{algorithm}[t]
	\caption{The Translate-Intent loss.}
	\begin{algorithmic}[1]
	\Let{$\mathsf{cls_{tar}}, \mathsf{tokens_{tar}}$}{$\mathsf{XLM}(\mathsf{x_{tar}})$}
	\Let{$\mathsf{pred_{ic}}$}{$\mathsf{IC}(\mathsf{cls_{tar}})$}
	\Let{$\mathsf{\mathcal{L}_{ti}}$}{$\mathsf{ce\_loss(pred_{ic}, y_{ic})}$}
	\Let{$\mathsf{\mathcal{L}_{total}}$}{$\mathsf{\mathcal{L}_{ic} + \mathcal{L}_{ec} + \mathcal{L}_{ti}}$}
    \end{algorithmic}
    \label{alg:translate}
\end{algorithm}

\subsection{Adaptive Weighting of Auxiliary Losses}
\label{adaptive}
In order to evaluate the benefits of combinations of two or more alignments, we employ the Multi-Loss Weighting with Coefficient of Variations \cite{groenendijk2020cov} technique (CoV) to calculate a weighted sum of auxiliary losses ($\mathsf{Aux}$) that we add to the main XNLU losses $\mathsf{\mathcal{L}_{ic} \normalfont{\ and\ } \mathcal{L}_{ec}}$ as follows:

\begin{equation}
\label{eq:weighted_sum}
	\mathsf{\mathcal{L}_{total} = \mathcal{L}_{ic} + \mathcal{L}_{ec} + \sum_{a \in Aux}w_{a}\mathcal{L}_{a}}
\end{equation}

The sole difference to CoV is that we opt to omit the loss weight normalisation step before application. The weights for an auxiliary loss $\mathsf{\mathcal{L}_{a,t}}$ for $\mathsf{a \in Aux}$ at training step $\mathsf{t}$ are calculated as follows:

\begin{equation}
	\mathsf{w_{a,t} = \frac{\sigma_{\ell_{a,t}}}{\mu_{\ell_{a,t}}} \ \ \ \ell_{a,t} = \frac{\mathcal{L}_{a,t}}{\mu_{\mathcal{L}_{a,t-1}}}}
\end{equation}

\noindent where $\mathsf{\ell_{a,t}}$ is the loss ratio of loss $\mathsf{a \in Aux}$ at training step $\mathsf{t}$, $\mathsf{\sigma}$ is the standard deviation over the history of loss ratios and $\mathsf{\mu_{\ell_{a,t-1}}}$ is the mean of the loss ratio $\mathsf{\ell_a}$ up to and including step $\mathsf{t-1}$. We also compare CoV to a simple sum of all losses i.e. equal weight for each loss, as shown in Algorithms \ref{alg:crossaligner}, \ref{alg:contrastive} and \ref{alg:translate} (line beginning with $\mathsf{\mathcal{L}_{total}}$).

\begin{table*}[t]
\centering
\setlength{\tabcolsep}{4pt}
    \begin{tabular}{l||ccc||c|c}
        \toprule
        \textbf{Models} & \textbf{MTOP (5)} & \textbf{MTOD (2)} & \textbf{M-ATIS (8)} & \textbf{MEAN (15)} & \textbf{Overall} \\\midrule
        
        Zero-Shot & 91.7/77.1 & 94.1/75.1 & 91.1/79.9 & 91.7/76.5 & 84.1 \\
        Target Language & 95.7/88.7 & 98.4/91.8 & 92.5/88.9 & 94.3/89.2 & 91.8 \\\midrule
        
        Translate-Train SOTA & 94.5/77.9 & 97.5/67.9 & 94.9/78.0 & 95.1/76.6 & 85.9 \\
        Translate-Intent (Ours) & 95.2/77.1 & 98.1/76.5 & 95.9/80.0 & 95.9/78.5 & 87.2 \\\midrule
        
        Previous SOTA & \textbf{95.6}/80.3 & \textbf{98.8}/72.9 & 96.0/81.2 & 96.1/79.8 & 88.0 \\
        XeroAlign$\mathsf{_{IO}}$ (Ours) & 95.3/81.3 & 98.5/75.1 & 96.4/82.3 & 96.3/81.1 & 88.7 \\
        CrossAligner (Ours) & 94.4/81.6 & 95.3/78.8 & 94.8/\textbf{84.1} & 94.7/\textbf{82.5} & 88.6 \\
        Contrastive (Ours) & 95.3/80.9 & 98.3/\textbf{79.6} & 96.5/79.3 & 96.3/79.8 & 88.1 \\
        XeroAlign$\mathsf{_{IO}}$ + CrossAligner (1+1) & 95.3/81.5 & 98.6/78.2 & 96.2/81.6 & 96.2/81.1 & 88.7 \\
        XeroAlign$\mathsf{_{IO}}$ + CrossAligner (CoV) & 95.4/\textbf{82.2} & \textbf{98.8}/78.3 & \textbf{96.6}/83.1 & \textbf{96.5}/82.1 & \textbf{89.3} \\
        \bottomrule
    \end{tabular}
\caption{Accuracy/F-Score for MTOP, MTOD, M-ATIS (number of non-English languages in brackets), MEAN over all datasets. Translate-Train SOTA is \cite{li2020mtop} for MTOP/MTOD and \cite{xu2020end} for M-ATIS.}
\label{tab:single_loss}
\end{table*}

\section{Experimental Setup}

\paragraph{Datasets} Three multilingual datasets are used to compare our methods with their most relevant counterparts. The datasets, which are used as standard benchmarks for the XNLU tasks, comprise nine unique languages (de, pt, zh, ja, hi, tr, fr, es, th) from 15 test sets (20,000+ instances in total) featuring diverse examples of users interacting with task-oriented personal assistants designed to test the XNLU capabilities of multilingual models. Two related tasks are being evaluated, Intent Classification and Entity/Slot Recognition.

\paragraph{Multilingual Task-Oriented Parsing} (MTOP) comprises 15K-22K utterances in each of 6 languages (en, de, fr, es, hi, th) spanning 11 domains \cite{li2020mtop}. The \textbf{Multilingual Task-Oriented Dialogue} (MTOD) consists of 43K English, 8K Spanish and 5K Thai utterances covering 3 domains \cite{schuster2018cross}. The \textbf{Multilingual ATIS++} (M-ATIS) contains up to 4.5K commands in each of 8 languages (en, es, pt, de, fr, zh, ja, hi, tr) featuring user interactions with a travel information system \cite{xu2020end}.

\paragraph{XLM}
Our pretrained language model of choice is XLM-RoBERTa \cite{conneau2019unsupervised}. We use the large (550M parameters) model from HuggingFace \cite{wolf2019huggingface} with a $\mathsf{hidden\_size}$ = 1,024. 

\paragraph{Training Setup}
We use a minimalist setup that features default settings and components to focus the results on the methods rather than hyperparameter tuning or custom architecture design. We implemented all models with PyTorch using fixed hyperparameters between experiments except for MTOD, where due to its size, we trained with fewer epochs and a lower learning rate (both 50\% lower\footnote{Download code and data at \url{https://github.com/huawei-noah/noah-research}}).

\section{Results}
\label{reults}

\paragraph{Terminology} Henceforth, we refer to models trained with labelled data in each language as \textbf{Target Language}, the models trained only on English data as \textbf{Zero-Shot}, our translate-intent method as \textbf{Translate-Intent} (\textbf{TI}), the scores reported by \citet{gritta2021xeroalign} as \textbf{Previous SOTA}, our IO-only implementation of that model as \textbf{XeroAlign$\mathsf{_{IO}}$} (\textbf{XA}$\mathsf{_{IO}}$), our contrastive alignment method as \textbf{Contrastive} (\textbf{CTR}) and our main method as \textbf{CrossAligner} (\textbf{CA}). Lastly, the simple sum of alignment losses is referred to as \textbf{1+1} and the weighted sum from \ref{adaptive} as \textbf{CoV}.

\paragraph{Metrics}
We use Accuracy for intent classification and F-Score for entity/slot recognition. In addition, we use an Overall score (the average of F-Score and Accuracy) for model ranking, similar to \citet{hu2020xtreme,wang2019superglue,wang2018glue}. Results are shown as averages (MEAN) over all test sets and datasets, presented in Tables \ref{tab:single_loss} and \ref{tab:multi_loss}. Intent classification is thus evaluated on $\sim$20,000 diverse user commands and entity recognition on $\sim$60,000 individual slots from 100+ slot types. For a full breakdown, see Tables \ref{tab:multi_atis}, \ref{tab:mtop} and \ref{tab:multi_loss_full} in Appendix \ref{sec:appendix_tables}.

\paragraph{Statistical Significance}
For a robust comparison with the previous SOTA, we conduct a two-tailed z-test for the difference of proportions \cite{schumacker2017z}. Our most effective method is statistically significant for all datasets at p < 0.01. The margin of improvement for slot tagging is +2.3 (F-Score) over previous SOTA and significant at p < 0.01.

\subsection{Individual Zero-Shot Transfer Methods}

\begin{table*}[ht]
	\centering
	\setlength{\tabcolsep}{4pt}
	\begin{tabular}{c||cccc||cc|c||cc|c}
		\toprule
		\multirow{2}{*}{\textbf{Setup}} & \multicolumn{4}{c||}{\textbf{Auxiliary Losses}} & \multicolumn{3}{c||}{\textbf{CoV Weighting}} & \multicolumn{3}{c}{\textbf{1+1 Weighting}} \\
		\cline{2-11}
		& \rule{0pt}{2.5ex} \textbf{CA} & \textbf{XA}$_\mathsf{IO}$ & \textbf{CTR} & \textbf{TI} & \multicolumn{2}{c|}{\textbf{MEAN (15)}} & \textbf{Overall} & \multicolumn{2}{c|}{\textbf{MEAN (15)}} & \textbf{Overall} \\
		\midrule
		\multirow{9}{*}{\textbf{2-Loss}}
		& x & x &  &  & \textbf{\scimean{96.5}} & \textbf{\scsmean{82.1}} & \textbf{\scovral{89.3}} & \scimean{96.2} & \scsmean{81.1} & \scovral{88.7} \\
		&  & x & x &  & \scimean{95.9} & \scsmean{80.1} & \scovral{88.0} & \scimean{96.1} & \scsmean{80.1} & \scovral{88.1} \\
		& x &  & x &  & \scimean{96.2} & \scsmean{81.3} & \scovral{88.8} & \scimean{96.1} & \scsmean{78.2} & \scovral{87.2} \\
		& x &  &  & x & \scimean{96.2} & \scsmean{81.3} & \scovral{88.8} & \scimean{96.2} & \scsmean{79.2} & \scovral{87.7} \\
		&  & x &  & x & \scimean{96.2} & \scsmean{80.3} & \scovral{88.3} & \scimean{96.3} & \scsmean{80.2} & \scovral{88.3} \\
		&  &  & x & x & \scimean{96.1} & \scsmean{79.6} & \scovral{87.9} & \scimean{96.2} & \scsmean{79.7} & \scovral{88.0} \\
		\midrule
		\multirow{6}{*}{\textbf{3-Loss}}
		& x & x & x &  & \scimean{96.4} & \scsmean{81.4} & \scovral{88.9} & \scimean{96.3} & \scsmean{80.1} & \scovral{88.2} \\
		& x & x &  & x & \textbf{\scimean{96.5}} & \scsmean{80.6} & \scovral{88.6} & \scimean{96.2} & \scsmean{81.0} & \scovral{88.6} \\
        & x &  & x & x & \scimean{96.3} & \scsmean{81.2} & \scovral{88.8} & \scimean{96.3} & \scsmean{79.0} & \scovral{87.7} \\
		&  & x & x & x & \scimean{96.1} & \scsmean{80.3} & \scovral{88.2} & \scimean{96.4} & \scsmean{80.0} & \scovral{88.2} \\ 
		\midrule
		\textbf{4-Loss}
		& x & x & x & x & \scimean{96.3} & \scsmean{79.7} & \scovral{88.0} & \scimean{96.4} & \scsmean{79.7} & \scovral{88.1} \\
		\bottomrule
	\end{tabular}
	\caption{The Accuracy and F-Score for combinations of auxiliary losses with different weighting schemes. The number of non-English test languages is shown in brackets, MEAN is computed for all languages in the 3 XNLU datasets. More detailed breakdowns of each dataset and language are shown in Tables \ref{tab:multi_atis}, \ref{tab:mtop} and \ref{tab:multi_loss_full} in Appendix \ref{sec:appendix_tables}.}
	\label{tab:multi_loss}
\end{table*}

\paragraph{CrossAligner}
The focus of our primary method was to improve slot filling as the model must classify dozens of entity types in each dataset and to that end, it is an effective approach. CrossAligner exceeds the F-Score of the Previous SOTA by 2.7 points (82.5 versus 79.8). This is 1.4 points higher than XeroAlign$\mathsf{_{IO}}$ and 6 points higher than Zero-Shot. Despite the intent accuracy being 1.4 points lower than Previous SOTA and 1.6 lower than XeroAlign$\mathsf{_{IO}}$, 94.7 is still 0.4 higher than Target Language. CrossAligner's overall score is 0.6 higher than previous SOTA, which outperformed the common 'translate-train' models, including entity projection and word alignment. In order to demonstrate the necessity and specificity of the proposed architecture, we tested mean-pooled token embeddings as well as a $\mathsf{CLS}$ embedding as the input to CrossAligner instead of the entity classifier logits. The scores declined from 94.7/82.5 (88.6 Overall) to 92.3/80 (86.2 Overall) with a $\mathsf{CLS}$ sentence representation and 82.1/78.7 (80.4 Overall) for mean-pooled embeddings. Future applications of our method to other NLP tasks must note that CrossAligner is most effective for tasks with a \textit{complex entity tag set} where the presence of entities in a sentence is informative, i.e. a higher complexity and slot density should lead to a higher performance. In addition, CrossAligner combines well with other losses as we show in Section \ref{combos}. 

\paragraph{Translate-Intent}
Our alternative to the common `translate-train' baseline is not only conceptually simpler (no explicit slot recognition training), it also outperforms the previous Translate-Train SOTA scores (78.5 vs 76.6 F-Score, 95.9 vs 95.1 accuracy and 87.2 vs 85.9 Overall). Translate-Intent does not require error-prone preprocessing such as word/label alignment and can therefore be readily used as a default `translate-train' baseline in future work. Note that using mean-pooled token embeddings as sentence representations is not recommended for Translate-Intent as this causes the F-Score to decline sharply (-25 points).

\paragraph{Contrastive Alignment}
Despite orders of magnitude less data than used in related work (Section \ref{related}), our Contrastive Alignment showed a marginal improvement over the previous SOTA on intent classification (96.3 vs 96.1) thus by 0.1 Overall. That said, even though the contrastive loss pushes negative sentence embeddings away from the positives, this does not seem to confer a strong advantage over the previous SOTA, which only used positive examples. We have also evaluated an implementation of Contrastive Alignment using mean-pooled token embeddings as sentence representations, however, the Overall score declined to 86.8 (versus 88.1 with a standard $\mathsf{CLS}$ embedding).

\paragraph{XeroAlign$\mathsf{_{IO}}$}
Our implementation of the previous SOTA with an additional post-processing step (described in \ref{bio}) increased the F-Score by 1.3 points and accuracy by 0.2 (+0.7 Overall). For a comparison, training XeroAlign$\mathsf{_{IO}}$ with the conventional BIO tags results in a drop of 1.8 points (81.1 to 79.3 F-Score) on entity recognition and 0.4 on intent classification (96.1 to 95.7). Mean-pooled tokens are not recommended for XeroAlign$\mathsf{_{IO}}$ as this yields a 2-point decline (88.7 to 86.7 Overall). Other models also benefit from IO-only training, for example, the Zero-Shot model gains 2.6 points (73.9 up to 76.5 F-Score). One theoretical limitation of IO-only training is that given a sequence of `B-LOC I-LOC B-LOC', the IO-only models would incorrectly classify this as a single entity. However, in practice, this is rare and not something we have seen during preprocessing or error analysis. 

\subsection{Combinations of Losses}
\label{combos}
As our alignment methods have different strengths and weaknesses, we have also evaluated their combinations (see Table \ref{tab:multi_loss}) as either a simple sum of losses (\textbf{1+1}) or a weighted sum of losses (\textbf{CoV}) using the Coefficient of Variation. The highest overall score was achieved by a CoV-weighted combination of XeroAlign$\mathsf{_{IO}}$ and CrossAligner, which considerably improved on the previous SOTA (96.5 vs 96.1 Accuracy, 82.1 vs 79.8 F-Score, 89.3 vs 88.0 Overall). In total, three individual and almost a dozen combinations of losses improve over the best previously reported scores. In the following paragraphs, we analyse and explain why the combinations that include CrossAligner consistently produce higher scores and why adding more losses can result in diminishing returns.

\begin{figure}[t]
	\centering
	\includegraphics[width=\linewidth]{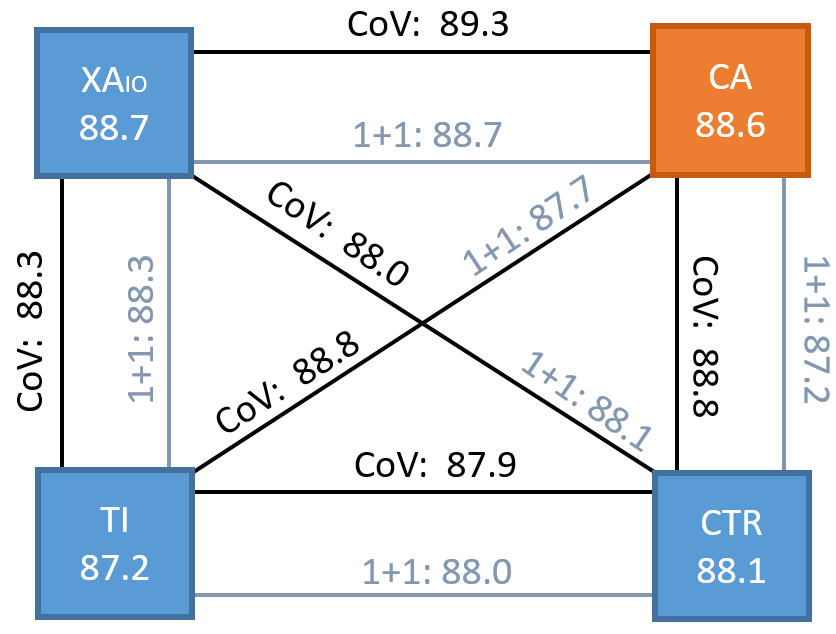}
	\caption{Overall scores for combinations of auxiliary losses weighted using either CoV or 1+1 (simple sum).}
	\label{fig:weighting}
\end{figure}

\paragraph{Compatibility of Losses}
We propose a hypothesis that can further help us interpret the numbers in Table \ref{tab:multi_loss}. It states that combining losses which use dissimilar sentence representations may be more beneficial than combining losses using similar sentence embeddings. In order to test that assumption, we clustered our alignment methods into two groups based on how their sentence representations are obtained: 1) XeroAlign$\mathsf{_{IO}}$, Translate-Intent and Contrastive Alignment, which all use the \textit{$\mathsf{CLS}$ embedding} and 2) CrossAligner, which aligns through the \textit{token embeddings} (used as the entity classifier input). In Figure \ref{fig:weighting}, we note that for combinations of any two alignment losses using the $\mathsf{CLS}$ embedding (shown as blue squares), there is no difference in the overall scores when using CoV or 1+1. However, when combining losses with different sentence representations (orange with any blue square) using CoV weighting, we observe consistent increases over the 1+1 setup (on average 1+ point overall) as well as an increase over their highest individual score. Additionally, in a 3-loss combination, we note that adding CrossAligner to any two losses from the $\mathsf{CLS}$-embedding group using CoV weighting yields an average improvement of 0.7 points compared to no improvement using 1+1.

\paragraph{Oversaturation of Losses} Another important observation harks back to our hypothesis stating that alignment methods with similar input embeddings do not lend themselves to being readily combined. We offer further evidence of this by testing combinations of CrossAligner with each of the $\mathsf{CLS}$-embedding losses [XA$\mathsf{_{IO}}$, TI and CTR], however, we use mean-pooled embeddings. The Overall scores decline in line with our hypothesis (XA$\mathsf{_{IO}}$ by -1.2, TI by -4.9, CTR by -0.6) with CoV-weighted losses and even more with the 1+1 weighted combinations (XA$\mathsf{_{IO}}$ by -2.1, TI by -7.6, CTR by -1.4). Similarly, combining multiple $\mathsf{CLS}$-embedding losses leads to a gradually diminishing benefit relative to the individual scores. Once again, the CoV-weighted losses show a significantly lower decline than the 1+1 combinations (Table \ref{tab:multi_loss}). Note that in our multi-loss scenario, intent classification remains unaffected by the choice of input embeddings as the accuracy remains stable at SOTA levels across experiments. We think this is due to the unequal task difficulty. In other words, sentence-level inference (intent recognition) is easier than token-level inference (entity recognition). 

\begin{table*}[t]
	\centering
	\setlength{\tabcolsep}{10pt}
	\begin{tabular}{l|ccccccc|c}
		\toprule \textbf{Categories} & \textbf{TH} & \textbf{HI} & \textbf{FR} & \textbf{DE} & \textbf{ZH} & \textbf{ES} & \textbf{PT} & \textbf{MEAN} \\ \midrule
		\textbf{Acceptable Transfer} & 28 & 51 & 53 & 40 & 37 & 38 & 23 & 38.6 \\
		\textbf{Partial Transfer} & 34 & 15 & 13 & 30 & 34 & 47 & 58 & 33.0 \\
		\textbf{Poor Transfer} & 38 & 34 & 34 & 30 & 29 & 15 & 19 & 28.4 \\ \midrule
		\textbf{Boundary Error} & 72 & 43 & 44 & 52 & 33 & 47 & 64 & 50.7 \\
		\textbf{Semantics Error} & 38 & 40 & 37 & 38 & 59 & 30 & 32 & 39.1  \\
		\textbf{Annotation Error} & 8 & 26 & 11 & 20 & 30 & 17 & 17 & 18.4 \\ \bottomrule
	\end{tabular}
	\caption{The summary of our qualitative error analysis with native speakers (700 samples from 7 languages).}
	\label{tab:errors}
\end{table*} 

\section{Error Analysis}
\label{error_analysis}

In order to contextualise the numbers reported in Tables \ref{tab:single_loss} and \ref{tab:multi_loss} in relevant linguistic insights, we have conducted a qualitative error analysis and present the highlights in this section. Readers interested in language-specific analysis (including many more examples) are encouraged to read \textbf{Appendix \ref{errors_per_language}}. We focused on errors committed by CrossAligner and XeroAlign$\mathsf{_{IO}}$, which achieved the best individual and combined scores. We sampled 100 random errors from each of the following settings: Hindi, French and German from MTOP, Portuguese, Chinese and Spanish from M-ATIS and Thai from MTOD for a diverse pool of errors. The authors adjudicated with native speakers to categorise mistakes into the following types.

\paragraph{Error Types}

We discovered two main sources of mistakes: A \textit{boundary error} occurs when the model predicts more or fewer entity words/tokens than given in the gold annotation. A \textit{semantics error} occurs when the wrong entity class/type is predicted. Models can therefore commit: 1) both errors resulting in \textit{Poor Transfer}, 2) a boundary error without a semantic error and vice versa giving us a \textit{Partial Transfer} or 3) neither error (a false negative), which we deemed an \textit{Acceptable Transfer}. We report individual and average error occurrences as well as transfer type percentages in Table \ref{tab:errors}.

\paragraph{Poor Transfer} indicates that the prediction error is too serious and unusable (even misleading) in a real-world personal assistant. This is typically due to both a boundary and a semantics error, however, some mistakes can be serious enough alone to result in poor transfer. For example, a boundary error can cause the retrieved name of a dish, person or a location to be incomplete and therefore invalid. A semantics error that classifies `10 secondes' (French) as `date\_time' instead of `music\_rewind\_time' would elicit the wrong agent response thus is unusable. On average, $\sim$28\% of mistakes fall into the `Poor Transfer' category.

\paragraph{Partial Transfer} is defined as either a boundary or a semantics error where neither is considered a serious problem. Such entities could be made usable in a personal assistant application with simple post-processing rules. Around 33 percent of errors were deemed to be partially correct. Often, this was due to including some adjacent punctuation or an article/preposition as part of the entity or a slightly shorter/longer news headline even though a search engine query with that string would have returned the relevant article. Entities such as `24 minat ka' (Hindi) versus `24 minat' (24 minutes) exemplify the fact that a disputed entity boundary is the most frequent source of error in this category. On the semantics side, we considered a location partially correct if `state\_name' instead of `city\_name' (for Washington D.C.) was predicted, a location was expected and the boundary was accurate. 
\paragraph{Acceptable Transfer} These examples are `errors' we considered correct and usable `as is' because neither the entity boundary nor its semantics were thought to be wrong. On average, we deemed almost 39\% of entities acceptable for a real-world personal assistant application with around half of those being down to \textit{annotation problems} (labels missing or incorrect). In other cases, we accepted predictions that offered a valid alternative e.g. when both `me' (French) and `je' (I/me) are present in the user utterance and both refer to the same `person\_reminded'. Valid alternatives were predicted but annotated somewhat differently. For example, when the entity boundary was slightly wider `de ida e volta' (Portuguese) instead of `ida e volta' (round trip) where both entities are correct. Similarly, classifying `salmon' as an ingredient rather than a dish (when `salmon' is an object of `prepare') was considered an acceptable transfer.

\subsection{Error Analysis Summary}

While the intent classification task is transferred well in a cross-lingual setting, performing better than training on labelled data, our SOTA slot recognition F-Score is almost 7 points behind Target Language. We think there are several factors involved. Articles, some prepositions, conjunctions, determiners and/or possessives do not transfer easily and may largely be ignored by the XLM as they don't carry important sentence level (e.g. intent) semantics. English is not ideal as a cross-lingual pivot for many of the dozens of languages covered by the XLM as elements of culture and vernacular that may not have a direct English equivalent don't transfer easily in a zero-shot setting. Aligning on the most well-resourced language in the same family should help \cite{xia2021metaxl}. The limits of machine translation, especially for low-resource languages \cite{mager2021findings}, can further inhibit alignment methods that leverage parallel data. Inconsistency of annotation (intra-language and inter-language) is a source of errors when the key concepts are learnt in one language and evaluated (sometimes unreliably) in the target language. Finally, there were no substantial qualitative differences between XeroAlign$\mathsf{_{IO}}$ and CrossAligner in our error analysis suggesting that the aforementioned error patterns may be a feature of the XLM model itself, the nature of the datasets or some as yet unknown confounding variable rather than the choice of the alignment method.

\section{Conclusions}

We have introduced a variety of cross-lingual methods for task-oriented XNLU to enable effective zero-shot transfer by learning alignment with unlabelled parallel data. The principal method, \textit{CrossAligner}, transforms English train data into a new language-agnostic task used to align model predictions across languages, achieving SOTA on entity recognition. We then presented a \textit{Contrastive Alignment} that optimises for a small cosine distance between translated sentences while increasing it between unrelated sentences, using orders of magnitude less data than previous works. We proposed \textit{Translate-Intent}, a fast and simple baseline that beats previous Translate-Train SOTA approaches without error-prone data transformations such as slot label projection. The best overall performance across nine languages, fifteen tests sets and three task-oriented multilingual datasets was achieved by a \textit{Coefficient of Variation} weighted combination of CrossAligner and XeroAlign$\mathsf{_{IO}}$. Our quantitative analysis investigated which types of auxiliary losses yield the most effective combinations. This resulted in several proposed configurations also exceeding previous SOTA scores. Our detailed qualitative error analysis revealed that the best methods have the potential to approach target language performance as most errors were deemed to be of low to medium severity. We hope our contributions and resources will inspire exciting future work in this fascinating NLP research area.

\section*{Acknowledgements}
We want to thank Philip John Gorinski, Guchun Zhang, Sushmit Bhattacharjee and Nicholas Aussel for providing the native speaker expertise for our qualitative error analysis. We are grateful to the ARR reviewers for their insightful comments and feedback. We also want to thank the MindSpore\footnote{\url{https://github.com/mindspore-ai}}\footnote{\url{https://mindspore.cn/}} team members for the technical support.

\bibliography{anthology,custom}

\begin{thebibliography}{36}
\expandafter\ifx\csname natexlab\endcsname\relax\def\natexlab#1{#1}\fi

\bibitem[{Arivazhagan et~al.(2019)Arivazhagan, Bapna, Firat, Aharoni, Johnson,
  and Macherey}]{arivazhagan2019missing}
Naveen Arivazhagan, Ankur Bapna, Orhan Firat, Roee Aharoni, Melvin Johnson, and
  Wolfgang Macherey. 2019.
\newblock The missing ingredient in zero-shot neural machine translation.
\newblock \emph{arXiv preprint arXiv:1903.07091}.

\bibitem[{Artetxe et~al.(2020)Artetxe, Ruder, and Yogatama}]{artetxe2019cross}
Mikel Artetxe, Sebastian Ruder, and Dani Yogatama. 2020.
\newblock On the cross-lingual transferability of monolingual representations.
\newblock In \emph{Proceedings of the 58th Annual Meeting of the Association
  for Computational Linguistics}, pages 4623--4637.

\bibitem[{Artetxe and Schwenk(2019)}]{artetxe2019massively}
Mikel Artetxe and Holger Schwenk. 2019.
\newblock Massively multilingual sentence embeddings for zero-shot
  cross-lingual transfer and beyond.
\newblock \emph{Transactions of the Association for Computational Linguistics},
  7:597--610.

\bibitem[{Casanueva et~al.(2017)Casanueva, Budzianowski, Su, Mrk{\v{s}}i{\'c},
  Wen, Ultes, Rojas-Barahona, Young, and
  Ga{\v{s}}i{\'c}}]{casanueva2017benchmarking}
Inigo Casanueva, Pawe{\l} Budzianowski, Pei-Hao Su, Nikola Mrk{\v{s}}i{\'c},
  Tsung-Hsien Wen, Stefan Ultes, Lina Rojas-Barahona, Steve Young, and Milica
  Ga{\v{s}}i{\'c}. 2017.
\newblock A benchmarking environment for reinforcement learning based task
  oriented dialogue management.
\newblock \emph{arXiv preprint arXiv:1711.11023}.

\bibitem[{Chaudhary et~al.(2020)Chaudhary, Raman, Srinivasan, and
  Chen}]{chaudhary2020dict}
Aditi Chaudhary, Karthik Raman, Krishna Srinivasan, and Jiecao Chen. 2020.
\newblock Dict-mlm: Improved multilingual pre-training using bilingual
  dictionaries.
\newblock \emph{arXiv preprint arXiv:2010.12566}.

\bibitem[{Chen et~al.(2019)Chen, Zhuo, and Wang}]{chen2019bert}
Qian Chen, Zhu Zhuo, and Wen Wang. 2019.
\newblock Bert for joint intent classification and slot filling.
\newblock \emph{arXiv preprint arXiv:1902.10909}.

\bibitem[{Chi et~al.(2020)Chi, Dong, Wei, Yang, Singhal, Wang, Song, Mao,
  Huang, and Zhou}]{chi2020infoxlm}
Zewen Chi, Li~Dong, Furu Wei, Nan Yang, Saksham Singhal, Wenhui Wang, Xia Song,
  Xian-Ling Mao, Heyan Huang, and Ming Zhou. 2020.
\newblock Infoxlm: An information-theoretic framework for cross-lingual
  language model pre-training.
\newblock \emph{arXiv preprint arXiv:2007.07834}.

\bibitem[{Conneau et~al.(2020)Conneau, Khandelwal, Goyal, Chaudhary, Wenzek,
  Guzm{\'a}n, Grave, Ott, Zettlemoyer, and Stoyanov}]{conneau2019unsupervised}
Alexis Conneau, Kartikay Khandelwal, Naman Goyal, Vishrav Chaudhary, Guillaume
  Wenzek, Francisco Guzm{\'a}n, {\'E}douard Grave, Myle Ott, Luke Zettlemoyer,
  and Veselin Stoyanov. 2020.
\newblock Unsupervised cross-lingual representation learning at scale.
\newblock In \emph{Proceedings of the 58th Annual Meeting of the Association
  for Computational Linguistics}, pages 8440--8451.

\bibitem[{Devlin et~al.(2019)Devlin, Chang, Lee, and
  Toutanova}]{devlin2018bert}
Jacob Devlin, Ming-Wei Chang, Kenton Lee, and Kristina Toutanova. 2019.
\newblock Bert: Pre-training of deep bidirectional transformers for language
  understanding.
\newblock In \emph{Proceedings of the 2019 Conference of the North American
  Chapter of the Association for Computational Linguistics: Human Language
  Technologies, Volume 1 (Long and Short Papers)}, pages 4171--4186.

\bibitem[{Dou and Neubig(2021)}]{dou2021word}
Zi-Yi Dou and Graham Neubig. 2021.
\newblock Word alignment by fine-tuning embeddings on parallel corpora.
\newblock In \emph{Proceedings of the 16th Conference of the European Chapter
  of the Association for Computational Linguistics: Main Volume}, pages
  2112--2128.

\bibitem[{Dyer et~al.(2013)Dyer, Chahuneau, and Smith}]{dyer2013simple}
Chris Dyer, Victor Chahuneau, and Noah~A Smith. 2013.
\newblock A simple, fast, and effective reparameterization of ibm model 2.
\newblock In \emph{Proceedings of the 2013 Conference of the North American
  Chapter of the Association for Computational Linguistics: Human Language
  Technologies}, pages 644--648.

\bibitem[{Gritta and Iacobacci(2021)}]{gritta2021xeroalign}
Milan Gritta and Ignacio Iacobacci. 2021.
\newblock Xeroalign: Zero-shot cross-lingual transformer alignment.
\newblock \emph{arXiv:2105.02472}.

\bibitem[{Gritta et~al.(2021)Gritta, Lampouras, and
  Iacobacci}]{gritta2021conversation}
Milan Gritta, Gerasimos Lampouras, and Ignacio Iacobacci. 2021.
\newblock Conversation graph: Data augmentation, training, and evaluation for
  non-deterministic dialogue management.
\newblock \emph{Transactions of the Association for Computational Linguistics},
  9:36--52.

\bibitem[{Groenendijk et~al.(2021)Groenendijk, Karaoglu, Gevers, and
  Mensink}]{groenendijk2020cov}
Rick Groenendijk, Sezer Karaoglu, Theo Gevers, and Thomas Mensink. 2021.
\newblock Multi-loss weighting with coefficient of variations.
\newblock In \emph{Proceedings of the IEEE/CVF Winter Conference on
  Applications of Computer Vision}, pages 1469--1478.

\bibitem[{Hu et~al.(2020)Hu, Ruder, Siddhant, Neubig, Firat, and
  Johnson}]{hu2020xtreme}
Junjie Hu, Sebastian Ruder, Aditya Siddhant, Graham Neubig, Orhan Firat, and
  Melvin Johnson. 2020.
\newblock Xtreme: A massively multilingual multi-task benchmark for evaluating
  cross-lingual generalisation.
\newblock In \emph{International Conference on Machine Learning}, pages
  4411--4421. PMLR.

\bibitem[{Jain et~al.(2019)Jain, Paranjape, and Lipton}]{jain2019entity}
Alankar Jain, Bhargavi Paranjape, and Zachary~C. Lipton. 2019.
\newblock \href {https://doi.org/10.18653/v1/D19-1100} {Entity projection via
  machine translation for cross-lingual {NER}}.
\newblock In \emph{Proceedings of the 2019 Conference on Empirical Methods in
  Natural Language Processing and the 9th International Joint Conference on
  Natural Language Processing (EMNLP-IJCNLP)}, pages 1083--1092, Hong Kong,
  China. Association for Computational Linguistics.

\bibitem[{Kuwanto et~al.(2021)Kuwanto, Aky{\"u}rek, Tourni, Li, and
  Wijaya}]{kuwanto2021low}
Garry Kuwanto, Afra~Feyza Aky{\"u}rek, Isidora~Chara Tourni, Siyang Li, and
  Derry Wijaya. 2021.
\newblock Low-resource machine translation for low-resource languages:
  Leveraging comparable data, code-switching and compute resources.
\newblock \emph{arXiv preprint arXiv:2103.13272}.

\bibitem[{Lample and Conneau(2019)}]{lample2019cross}
Guillaume Lample and Alexis Conneau. 2019.
\newblock Cross-lingual language model pretraining.
\newblock \emph{arXiv preprint arXiv:1901.07291}.

\bibitem[{Li et~al.(2021{\natexlab{a}})Li, He, and Xu}]{li2021cross}
Bing Li, Yujie He, and Wenjin Xu. 2021{\natexlab{a}}.
\newblock Cross-lingual named entity recognition using parallel corpus: A new
  approach using xlm-roberta alignment.
\newblock \emph{arXiv preprint arXiv:2101.11112}.

\bibitem[{Li et~al.(2021{\natexlab{b}})Li, Arora, Chen, Gupta, Gupta, and
  Mehdad}]{li2020mtop}
Haoran Li, Abhinav Arora, Shuohui Chen, Anchit Gupta, Sonal Gupta, and Yashar
  Mehdad. 2021{\natexlab{b}}.
\newblock \href {https://www.aclweb.org/anthology/2021.eacl-main.257} {{MTOP}:
  A comprehensive multilingual task-oriented semantic parsing benchmark}.
\newblock In \emph{Proceedings of the 16th Conference of the European Chapter
  of the Association for Computational Linguistics: Main Volume}, pages
  2950--2962, Online. Association for Computational Linguistics.

\bibitem[{Liang et~al.(2020)Liang, Duan, Gong, Wu, Guo, Qi, Gong, Shou, Jiang,
  Cao et~al.}]{liang2020xglue}
Yaobo Liang, Nan Duan, Yeyun Gong, Ning Wu, Fenfei Guo, Weizhen Qi, Ming Gong,
  Linjun Shou, Daxin Jiang, Guihong Cao, et~al. 2020.
\newblock Xglue: A new benchmark datasetfor cross-lingual pre-training,
  understanding and generation.
\newblock In \emph{Proceedings of the 2020 Conference on Empirical Methods in
  Natural Language Processing (EMNLP)}, pages 6008--6018.

\bibitem[{Oord et~al.(2018)Oord, Li, and Vinyals}]{vandenoord2018rep}
Aaron van~den Oord, Yazhe Li, and Oriol Vinyals. 2018.
\newblock Representation learning with contrastive predictive coding.
\newblock \emph{arXiv preprint arXiv:1807.03748}.

\bibitem[{Pan et~al.(2020)Pan, Hang, Qi, Shah, Yu, and
  Potdar}]{pan2020multilingual}
Lin Pan, Chung-Wei Hang, Haode Qi, Abhishek Shah, Mo~Yu, and Saloni Potdar.
  2020.
\newblock Multilingual bert post-pretraining alignment.
\newblock \emph{arXiv preprint arXiv:2010.12547}.

\bibitem[{Qi and Du(2020)}]{qi2020translation}
Kunxun Qi and Jianfeng Du. 2020.
\newblock Translation-based matching adversarial network for cross-lingual
  natural language inference.
\newblock In \emph{Proceedings of the AAAI Conference on Artificial
  Intelligence}, volume~34, pages 8632--8639.

\bibitem[{Qin et~al.(2020)Qin, Ni, Zhang, and Che}]{qin2020cosda}
Libo Qin, Minheng Ni, Yue Zhang, and Wanxiang Che. 2020.
\newblock Cosda-ml: Multi-lingual code-switching data augmentation for
  zero-shot cross-lingual nlp.
\newblock \emph{arXiv preprint arXiv:2006.06402}.

\bibitem[{Razumovskaia et~al.(2021)Razumovskaia, Glava{\v{s}}, Majewska,
  Korhonen, and Vuli{\'c}}]{razumovskaia2021crossing}
Evgeniia Razumovskaia, Goran Glava{\v{s}}, Olga Majewska, Anna Korhonen, and
  Ivan Vuli{\'c}. 2021.
\newblock Crossing the conversational chasm: A primer on multilingual
  task-oriented dialogue systems.
\newblock \emph{arXiv preprint arXiv:2104.08570}.

\bibitem[{Sang and De~Meulder(2003)}]{sang2003introduction}
Erik Tjong~Kim Sang and Fien De~Meulder. 2003.
\newblock Introduction to the conll-2003 shared task: Language-independent
  named entity recognition.
\newblock In \emph{Proceedings of the Seventh Conference on Natural Language
  Learning at HLT-NAACL 2003}, pages 142--147.

\bibitem[{Schuster et~al.(2019)Schuster, Gupta, Shah, and
  Lewis}]{schuster2018cross}
Sebastian Schuster, Sonal Gupta, Rushin Shah, and Mike Lewis. 2019.
\newblock Cross-lingual transfer learning for multilingual task oriented
  dialog.
\newblock In \emph{Proceedings of the 2019 Conference of the North American
  Chapter of the Association for Computational Linguistics: Human Language
  Technologies, Volume 1 (Long and Short Papers)}, pages 3795--3805.

\bibitem[{Siddhant et~al.(2020)Siddhant, Johnson, Tsai, Ari, Riesa, Bapna,
  Firat, and Raman}]{siddhant2020evaluating}
Aditya Siddhant, Melvin Johnson, Henry Tsai, Naveen Ari, Jason Riesa, Ankur
  Bapna, Orhan Firat, and Karthik Raman. 2020.
\newblock Evaluating the cross-lingual effectiveness of massively multilingual
  neural machine translation.
\newblock In \emph{Proceedings of the AAAI Conference on Artificial
  Intelligence}, volume~34, pages 8854--8861.

\bibitem[{Wang et~al.(2019)Wang, Pruksachatkun, Nangia, Singh, Michael, Hill,
  Levy, and Bowman}]{wang2019superglue}
Alex Wang, Yada Pruksachatkun, Nikita Nangia, Amanpreet Singh, Julian Michael,
  Felix Hill, Omer Levy, and Samuel~R Bowman. 2019.
\newblock Superglue: A stickier benchmark for general-purpose language
  understanding systems.
\newblock \emph{Advances in Neural Information Processing Systems}, 32.

\bibitem[{Wang et~al.(2018)Wang, Singh, Michael, Hill, Levy, and
  Bowman}]{wang2018glue}
Alex Wang, Amanpreet Singh, Julian Michael, Felix Hill, Omer Levy, and Samuel~R
  Bowman. 2018.
\newblock Glue: A multi-task benchmark and analysis platform for natural
  language understanding.
\newblock \emph{arXiv preprint arXiv:1804.07461}.

\bibitem[{Wolf et~al.(2019)Wolf, Debut, Sanh, Chaumond, Delangue, Moi, Cistac,
  Rault, Louf, Funtowicz et~al.}]{wolf2019huggingface}
Thomas Wolf, Lysandre Debut, Victor Sanh, Julien Chaumond, Clement Delangue,
  Anthony Moi, Pierric Cistac, Tim Rault, R{\'e}mi Louf, Morgan Funtowicz,
  et~al. 2019.
\newblock Huggingface's transformers: State-of-the-art natural language
  processing.
\newblock \emph{ArXiv}, pages arXiv--1910.

\bibitem[{Wu et~al.(2021)Wu, Wu, and Huang}]{wu2021rethink}
Chuhan Wu, Fangzhao Wu, and Yongfeng Huang. 2021.
\newblock Rethinking infonce: How many negative samples do you need?
\newblock \emph{arXiv preprint arXiv:2105.13003}.

\bibitem[{Xia et~al.(2021)Xia, Zheng, Mukherjee, Shokouhi, Neubig, and
  Awadallah}]{xia2021metaxl}
Mengzhou Xia, Guoqing Zheng, Subhabrata Mukherjee, Milad Shokouhi, Graham
  Neubig, and Ahmed~Hassan Awadallah. 2021.
\newblock Metaxl: Meta representation transformation for low-resource
  cross-lingual learning.
\newblock \emph{arXiv preprint arXiv:2104.07908}.

\bibitem[{Xu et~al.(2020)Xu, Haider, and Mansour}]{xu2020end}
Weijia Xu, Batool Haider, and Saab Mansour. 2020.
\newblock End-to-end slot alignment and recognition for cross-lingual nlu.
\newblock In \emph{Proceedings of the 2020 Conference on Empirical Methods in
  Natural Language Processing (EMNLP)}, pages 5052--5063.

\bibitem[{Yi and Cheng(2021)}]{yi2021zero}
Huixiong Yi and Jin Cheng. 2021.
\newblock Zero-shot entity recognition via multi-source projection and
  unlabeled data.
\newblock In \emph{IOP Conference Series: Earth and Environmental Science},
  volume 693, page 012084. IOP Publishing.

\end{thebibliography}
\bibliographystyle{acl_natbib}

\newpage
\appendix

\section{Appendix}
\label{sec:appendix}

\subsection{Error Analysis per Language}
\label{errors_per_language}

\paragraph{German} examples of partial transfer are boundary errors such as tagging punctuation `-' as part of the entity (e.g. in a timer or alarm name) as well as \textit{not} tagging some punctuation e.g. in `14. Mai' where `.' is equivalent to the English `th' in `14th' and is expected in German dates. Such entities can be used with a basic post-processing rule as their classes and contents were sufficiently well predicted. Similarly, including `die', `den', `das', `mir', `des', `der' in the retrieved entity, particularly in \textit{free-format entities} such as news headlines, text message contents and memos need not invalidate the prediction, e.g. `\textit{die} hausschliessung' (house closure), `\textit{des} Reiseverbots' (\textit{of} travel ban) and `\textit{den} Termin' (appointment). Just a few of such linguistic `bad habits' can quickly accumulate to cause more than half of all errors.

\paragraph{Chinese} cross-lingual transfer problems often include boundary issues featuring the `of' preposition or the possessive `'s' (\textit{de} in Chinese) e.g. `Zuì piányí \textit{de}' (cheapest), `Jiāzhōu \textit{de}' (California) or `āndàlüè \textit{de} hángbān' (flights to Ontario). Depending on context, we considered these at least partially correct rather than a failed transfer. More serious though less explicable errors were `Gěi wǒ zhōu'èr' (\textit{give me} Tuesday's), `Liè chū zhōu liù' (\textit{list} Saturday's) or `Xiǎnshì zhōusān' (\textit{show me }Wednesday's) where `give me', `list' and `show me' were tagged as part of `date\_time' a total of 20 times. The most frequent semantic (partial) error was `Washington D.C.', which was tagged as a state rather than a city no fewer than 16 times.

\paragraph{French} instances of acceptable transfer include tagging `Ankara' and `Turquie' separately rather than as a single chunk `Ankara, en Turquie' (possible annotation problem). Reminiscent of the patterns seen in other languages, articles tend to feature prominently in boundary errors, e.g. `\textit{la} famille' (family), `\textit{l'}arrosage' (watering), `\textit{les} elections' (elections), `\textit{le} chat Zoom' (Zoom chat) and `\textit{la} mere de Kylie' (Kylie's mother), which we considered usable `as is'. For an example of annotation inconsistency across languages, consider the entity `gros titres' or `top headlines' in English. The model correctly transferred the English tags for `top' (news\_reference) and `headline' (news\_type) although the French annotation was given as `gros titres' (news\_type), which is plausible but less coherent than the model's prediction. 

\paragraph{Spanish} Once again, articles, some prepositions and occasionally conjunctions e.g. `de', `las', `y', `la', `por' (of, the, and, a, in/for) have caused the majority of boundary errors, most of which are partially acceptable. Examples include time `\textit{las} 10 a.m.' (10 a.m.) and `no mas tarde que \textit{las}' (no later than), journey specifications `\textit{de} ida' and `ida \textit{y} vuelta' (round trip), periods of day `\textit{la} mañana' (morning) and `\textit{la} noche' (night), dates `seis \textit{de} junio' (June 6th) as well as skipping `\textit{en punto}' (o'clock) in `antes de las 4 p.m. \textit{en punto}' (before 4 p.m.). These minor errors show that the current SOTA in cross-lingual zero-shot transfer is close to solving these cases. Other errors such as `conexión' instead of `\textit{con} conexión' (\textit{with} connection) and `\textit{la} mañana' rather than `\textit{por la} mañana' (\textit{in} the morning) are more examples of disputed annotations. 

\paragraph{Portuguese} predictions closely follow Spanish error patterns and reflect the wider issues with articles and prepositions, e.g. `terça-feira \textit{de} manhã' (Tuesday morning), `\textit{de} ida e volta' (round trip), `cinco \textit{de} abril' (April 5th) or `5 horas \textit{da} tarde' (5 p.m.) with '\textit{de}' and `\textit{da}' (both \textit{of}) being the unannotated parts that did not transfer optimally. Other boundary mistakes were caused by `\textit{das}' (of) and `\textit{as}' (the), for example, `antes \textit{das} 6 horas da tarde' and `após \textit{as} 6 horas da tarde' (before and after 6 p.m). Annotations that needlessly punished cross-lingual transfer included `\textit{somente} de ida' (one way) where `\textit{somente}' (only) was not annotated in the English dataset and `econômica' (economy), which was annotated as `class\_type' in English, correctly transferred but flagged as wrong.

\paragraph{Hindi} errors have a relatively high number (26) of problematic annotations although most mistakes are caused by the now familiar improper handling of prepositions, articles and/or possessives e.g. `\textit{ke}', `\textit{tak}', `\textit{ka}' (`of', `by', `'s') in phrases such as '30 minat \textit{ka}' (30 minutes), `kitanee der \textit{tak}' (how long), `kal \textit{ke}' (yesterday's), `aaj \textit{ke}' (today's) or `1 baje \textit{ka}' (1 p.m.). Transliterated entities i.e. English pronunciation written in Devanagari, is the second largest category of transfer problems in Hindi, e.g. `pakrino romaano' (Pecorino Romano), `goda cheez' (Goda cheese), `braun aaid garl' (Brown Eyed Girl), `paindora' (Pandora), `pool leeg' (Pool League), `daayanaasor jooniyar' (Dinosaur Junior) or `da most byooteephul moment' (The Most Beatiful Moment). These are problematic because such entities are neither native to Hindi nor are they written in Latin alphabet hence may not have been observed in this form during XLM pretraining.

\paragraph{Thai} errors were analysed with a translation service as we were unable to secure a native speaker. Even so, we observed boundary errors previously seen in other languages. Words such as `\textit{nai}' (`of' or `in', the most frequent cause) and `\textit{bpai}' (`in', `off' or `to', no direct English translation) were the typical sources of boundary issues, e.g. '\textit{nai} sùt sàp-daa née' (this weekend), `\textit{nai} wan pút' (Wednesday), `\textit{bpai} séu XYZ' (go buy XYZ) and `\textit{nai} wan née' (today or on this day). Such patterns accounted for more than half of all mistakes. Machine translation can also be a source of errors. For example, the word `reminder' is an entity in English (tag: reminder/noun). It was translated as `kam dteuuan', however, `reminder' appears in Thai data as `dteuuan kwaam jam', which the model repeatedly missed, leading to 18 errors for what should be an easy case of zero-shot transfer.

	\begin{table*}[h]
		\centering
		\small
		\setlength{\tabcolsep}{4.5pt}
		\begin{tabular}{l|cccccccc|c}
			\toprule \textbf{Model} & \textbf{DE} & \textbf{ES} & \textbf{FR} & \textbf{TR} & \textbf{HI} & \textbf{ZH} & \textbf{PT} & \textbf{JA} & \textbf{MEAN} \\ \midrule
			Zero-Shot & 95.1/84.8 & 97.3/84.9 & 97.9/79.5 & 75.4/41.8 & 91.3/78.4 & 88.6/82.1 & 96.9/80.9 & 86.6/79.9 & 91.1/79.9 \\
			Target Language & 96.9/95.4 & 96.6/85.8 & 97.9/93.8 & 77.2/71.6 & 88.8/84.4 & 94.5/94.9 & 96.8/92.1 & 91.4/93.0 & 92.5/88.9 \\ 
			\midrule 
			Trans-Train SOTA & 96.7/89.0 & 97.2/76.4 & 97.5/79.6 & 93.7/61.7 & 92.8/78.6 & 96.0/83.3 & 96.8/76.3 & 88.3/79.1 & 94.9/78.0 \\ 
			Translate-Intent & 97.3/84.7 & 97.6/84.1 & 97.5/84.7 & 91.6/65.6 & 94.4/80.9 & 96.4/83.0 & 97.0/82.7 & 95.2/74.1 & 95.9/80.0 \\ \midrule
			Previous SOTA & \textbf{97.6}/84.9 & \textbf{97.8}/85.9 & 95.4/81.4 & 93.4/70.6 & 94.0/79.7 & 96.4/83.3 & 97.6/79.9 & 96.1/83.5 & 96.0/81.2 \\ 
			XeroAlign$\mathsf{_{IO}}$ & 97.4/84.1 & 97.4/\textbf{86.2} & 97.9/83.3 & 93.6/76.0 & \textbf{95.1}/80.1 & 96.0/83.7 & \textbf{98.0}/81.6 & 95.6/83.7 & 96.4/82.3 \\
			CrossAligner & 96.9/\textbf{90.4} & 97.4/73.1 & \textbf{98.0}/88.7 & 88.7/75.4 & 94.5/\textbf{86.6} & 94.2/88.2 & 97.4/\textbf{81.7} & 91.6/88.7 & 94.8/\textbf{84.1} \\
			Contrastive & 97.5/79.6 & 97.3/77.1 & 97.6/76.1 & 93.3/\textbf{76.3} & 94.6/79.8 & \textbf{96.9}/87.5 & 97.4/77.0 & \textbf{97.3}/80.9 & 96.5/79.3 \\
			\midrule
			XA$_\mathsf{IO}$, CA (1+1) & 97.3/89.7 & 97.5/72.3 & 97.8/82.1 & \textbf{94.0}/69.1 & 95.3/79.1 & 95.9/89.6 & 97.4/80.1 & 94.0/\textbf{91.1} & 96.2/81.6 \\
			XA$_\mathsf{IO}$, CA (CoV) & \textbf{97.6}/88.7 & 97.6/72.5 & \textbf{98.0}/\textbf{88.8} & 93.3/73.9 & \textbf{95.1}/79.7 & 96.5/\textbf{89.9} & 97.5/80.9 & 97.0/90.5 & \textbf{96.6}/83.1 \\
			\bottomrule
		\end{tabular}
		\caption{Accuracy/F-Score for M-ATIS. \newcite{xu2020end} is the previous translate-train SOTA.}
		\label{tab:multi_atis}
	\end{table*}
	
	\begin{table*}[h]
		\centering
		\small
		\setlength{\tabcolsep}{4.3pt}
		\begin{tabular}{l|ccccc|c||cc|c}
			\toprule \textbf{Model} & \textbf{DE} & \textbf{ES} & \textbf{FR} & \textbf{TH} & \textbf{HI} & \textbf{MEAN}  & \textbf{ES} & \textbf{TH} & \textbf{MEAN} \\ \midrule
			Zero-Shot & 90.5/80.1 & 93.8/81.7 & 92.5/83.2 & 89.7/64.9 & 91.8/75.5 & 91.7/77.1 & 97.5/87.3 & 90.6/62.8 & 94.1/75.1 \\
			Target Language & 96.5/88.7 & 96.1/90.6 & 95.6/89.3 & 95.0/87.0 & 95.1/87.8 & 95.7/88.7 & 98.9/89.3 & 97.8/94.3 & 98.4/91.8 \\ 	 \midrule
			Trans-Train SOTA & 94.8/80.0 & 96.3/84.8 & 95.1/82.5 & 92.1/65.6 & 94.2/76.5 & 94.5/77.9  & 98.0/83.0 & 96.9/52.8 & 97.5/67.9 \\ 
			Transl.Intent & 96.4/83.4 & 96.2/73.5 & 95.5/84.5 & 92.9/68.4 & 94.9/75.6 & 95.2/77.1 & 99.2/87.5 & 96.9/65.4 & 98.1/76.5 \\\midrule
			Previous SOTA & \textbf{96.6}/84.4 & 96.5/83.3 & 95.7/84.5 & \textbf{94.1}/69.1 & 95.2/80.1 & \textbf{95.6}/80.3  & 99.2/88.4 & \textbf{98.4}/57.3 & \textbf{98.8}/72.9 \\
			XeroAlign$\mathsf{_{IO}}$ & 96.4/86.1 & 96.4/\textbf{84.4} & 95.4/\textbf{86.2} & 93.1/69.5 & 95.1/80.5 & 95.3/81.3 & \textbf{99.3}/88.8 & 97.6/62.0 & 98.5/75.4 \\
			CrossAligner & 95.4/86.0 & 95.1/81.9 & 94.5/84.9 & 92.6/73.9 & 94.3/\textbf{81.1} & 94.4/81.6 & 98.2/87.1 & 92.4/\textbf{70.4} & 95.3/78.8 \\
			Contrastive & 96.3/84.3 & 96.2/83.3 & 95.5/85.2 & 93.0/70.6 & \textbf{95.5}/80.9 & 95.3/80.9 & 99.2/89.0 & 97.3/70.1 & 98.3/\textbf{79.6} \\\midrule
			
			XA$_\mathsf{IO}$, CA (1+1) & 96.3/85.3 & \textbf{96.6}/83.1 & \textbf{95.9}/85.9 & 93.3/72.8 & 94.5/80.4 & 95.3/81.5 & 99.2/88.7 & 98.0/67.7 & 98.6/78.2 \\
			XA$_\mathsf{IO}$, CA (CoV) & 96.4/\textbf{86.6} & 96.1/83.8 & 95.8/85.7 & 93.3/\textbf{74.3} & 95.2/80.4 & 95.4/\textbf{82.2} & \textbf{99.3}/\textbf{89.3} & 98.2/67.2 & \textbf{98.8}/78.3 \\
			\bottomrule
		\end{tabular}
		\caption{Accuracy/F-Score for MTOP (left) and MTOD (right). \newcite{li2020mtop} is previous translate-train SOTA.}
		\label{tab:mtop}
	\end{table*}

\begin{table*}[ht]
	\centering
	\footnotesize
	\setlength{\tabcolsep}{4pt}
    \begin{tabular}{c||cccc||c||cc|cc|cc||cc|c}
		\toprule
		\multirow{2}{*}{\textbf{Setup}} & \multicolumn{4}{c||}{\textbf{Auxiliary Losses}} & \multirow{2}{*}{\textbf{Weight.}} & \multicolumn{2}{c|}{\multirow{2}{*}{\textbf{MTOP(5)}}} & \multicolumn{2}{c|}{\multirow{2}{*}{\textbf{MTOD(2)}}} & \multicolumn{2}{c||}{\multirow{2}{*}{\textbf{M-ATIS(8)}}} & \multicolumn{2}{c|}{\multirow{2}{*}{\textbf{MEAN(15)}}} & \multirow{2}{*}{\textbf{Overall}}\\
		\cline{2-5}
        & \rule{0pt}{2ex} \textbf{CA} & \textbf{XA}$_\mathsf{IO}$ & \textbf{CTR} & \textbf{TI} & & & & & & & & & & \\
        \midrule
        \multirow{18}{*}{\textbf{2-Loss}}
        & x & x &  &  & CoV & \cimtop{95.4}&\csmtop{82.2} & \textbf{\cimtod{98.8}}&\csmtod{78.3} & \textbf{\cimatis{96.6}}&\textbf{\csmatis{83.1}} & \textbf{\cimean{96.5}}&\textbf{\csmean{82.1}} & \textbf{\covral{89.3}} \\ 
		& x & x &  &  & 1+1 & \cimtop{95.3}&\csmtop{81.5} & \cimtod{98.6}&\csmtod{78.2} & \cimatis{96.2}&\csmatis{81.6} & \cimean{96.2}&\csmean{81.1} & \covral{88.7} \\
		&  & x & x &  & CoV & \cimtop{95.1}&\csmtop{80.9} & \cimtod{97.3}&\csmtod{73.5} & \cimatis{96.0}&\csmatis{81.3} & \cimean{95.9}&\csmean{80.1} & \covral{88.0} \\
		&  & x & x &  & 1+1 & \cimtop{95.2}&\csmtop{79.1} & \cimtod{97.1}&\csmtod{77.1} & \cimatis{96.3}&\csmatis{81.5} & \cimean{96.1}&\csmean{80.1} & \covral{88.1} \\
		& x &  & x &  & CoV & \cimtop{95.2}&\textbf{\csmtop{82.3}} & \cimtod{98.6}&\csmtod{77.6} & \cimatis{96.3}&\csmatis{81.6} & \cimean{96.2}&\csmean{81.3} & \covral{88.8} \\
		& x &  & x &  & 1+1 & \cimtop{95.2}&\csmtop{82.2} & \cimtod{98.6}&\csmtod{78.3} & \cimatis{96.1}&\csmatis{75.7} & \cimean{96.1}&\csmean{78.2} & \covral{87.2} \\
		& x &  &  & x & CoV & \cimtop{95.3}&\csmtop{81.5} & \textbf{\cimtod{98.8}}&\csmtod{78.0} & \cimatis{96.1}&\csmatis{82.0} & \cimean{96.2}&\csmean{81.3} & \covral{88.8} \\
		& x &  &  & x & 1+1 & \textbf{\cimtop{95.5}}&\csmtop{81.3} & \cimtod{98.6}&\csmtod{77.4} & \cimatis{96.1}&\csmatis{78.3} & \cimean{96.2}&\csmean{79.2} & \covral{87.7} \\
		&  & x &  & x & CoV & \cimtop{95.2}&\csmtop{80.6} & \cimtod{98.5}&\csmtod{78.2} & \cimatis{96.2}&\csmatis{80.6} & \cimean{96.2}&\csmean{80.3} & \covral{88.3} \\
		&  & x &  & x & 1+1 & \cimtop{95.3}&\csmtop{79.3} & \textbf{\cimtod{98.8}}&\csmtod{78.1} & \cimatis{96.4}&\csmatis{81.3} & \cimean{96.3}&\csmean{80.2} & \covral{88.3} \\
		&  &  & x & x & CoV & \cimtop{95.2}&\csmtop{79.8} & \cimtod{97.8}&\csmtod{76.8} & \cimatis{96.2}&\csmatis{80.3} & \cimean{96.1}&\csmean{79.6} & \covral{87.9} \\
		&  &  & x & x & 1+1 & \cimtop{95.3}&\csmtop{79.9} & \cimtod{98.6}&\csmtod{79.2} & \cimatis{96.1}&\csmatis{79.7} & \cimean{96.2}&\csmean{79.7} & \covral{88.0} \\ 
        \midrule
		\multirow{12}{*}{\textbf{3-Loss}} & x & x & x &  & CoV & \cimtop{95.3}&\csmtop{82.1} & \cimtod{98.6}&\csmtod{78.7} & \textbf{\cimatis{96.6}}&\csmatis{81.7} & \cimean{96.4}&\csmean{81.4} & \covral{88.9} \\
		& x & x & x &  & 1+1 & \cimtop{95.3}&\csmtop{81.8} & \textbf{\cimtod{98.8}}&\csmtod{79.4} & \cimatis{96.4}&\csmatis{79.3} & \cimean{96.3}&\csmean{80.1} & \covral{88.2} \\
		& x & x &  & x & CoV & \cimtop{95.4}&\csmtop{81.4} & \cimtod{98.7}&\csmtod{78.4} & \textbf{\cimatis{96.6}}&\csmatis{80.6} & \textbf{\cimean{96.5}}&\csmean{80.6} & \covral{88.6} \\
		& x & x &  & x & 1+1 & \cimtop{95.4}&\csmtop{81.0} & \cimtod{98.0}&\csmtod{78.9} & \cimatis{96.4}&\csmatis{81.5} & \cimean{96.2}&\csmean{81.0} & \covral{88.6} \\
		& x &  & x & x & CoV & \cimtop{95.1}&\csmtop{81.5} & \textbf{\cimtod{98.8}}&\csmtod{79.2} & \cimatis{96.5}&\csmatis{81.4} & \cimean{96.3}&\csmean{81.2} & \covral{88.8} \\
		& x &  & x & x & 1+1 & \textbf{\cimtop{95.5}}&\csmtop{80.1} & \cimtod{98.7}&\csmtod{78.5} & \cimatis{96.1}&\csmatis{78.5} & \cimean{96.3}&\csmean{79.0} & \covral{87.7} \\
		&  & x & x & x & CoV & \cimtop{95.2}&\csmtop{80.2} & \cimtod{97.7}&\csmtod{78.4} & \cimatis{96.2}&\csmatis{80.8} & \cimean{96.1}&\csmean{80.3} & \covral{88.2} \\
		&  & x & x & x & 1+1 & \cimtop{95.2}&\csmtop{80.0} & \cimtod{98.5}&\csmtod{78.6} & \textbf{\cimatis{96.6}}&\csmatis{80.4} & \cimean{96.4}&\csmean{80.0} & \covral{88.2} \\
		\midrule
		\multirow{3}{*}{\textbf{4-Loss}} & x & x & x & x & CoV & \cimtop{95.3}&\csmtop{81.6} & \cimtod{98.7}&\textbf{\csmtod{79.7}} & \cimatis{96.3}&\csmatis{78.6} & \cimean{96.3}&\csmean{79.7} & \covral{88.0} \\
		& x & x & x & x & 1+1 & \cimtop{95.4}&\csmtop{81.1} & \cimtod{98.4}&\csmtod{79.1} & \textbf{\cimatis{96.6}}&\csmatis{78.9} & \cimean{96.4}&\csmean{79.7} & \covral{88.1} \\
		\bottomrule
	\end{tabular}
	\caption{Accuracy and F-Score for combinations of auxiliary losses with different weighting schemes.}
	\label{tab:multi_loss_full}
\end{table*}

\subsection{Full Tables}
\label{sec:appendix_tables}

The full language breakdown for MultiATIS++ (Table \ref{tab:multi_atis}) and MTOP+MTOD (Table \ref{tab:mtop}). Table \ref{tab:multi_loss_full} shows the full details of the combinations of losses from Table \ref{tab:multi_loss} in Results (\ref{reults}).

\end{document}